\documentclass[10pt,twocolumn,letterpaper]{article}

\usepackage{iccv}
\usepackage{times}
\usepackage{epsfig}
\usepackage{graphicx}
\usepackage{amsmath}
\usepackage{amssymb}
\usepackage{booktabs}
\usepackage{multirow}
\usepackage{cooltooltips}
\usepackage[table,xcdraw]{xcolor}

\usepackage[pagebackref=true,breaklinks=true,letterpaper=true,colorlinks,bookmarks=false]{hyperref}

 \iccvfinalcopy 

\def\iccvPaperID{9929} 

\ificcvfinal\fi

\begin{document}

\title{PRIOR: Prototype Representation Joint Learning \\ from Medical Images and Reports}

\author{Pujin Cheng \textsuperscript{1,2}, \quad Li Lin \textsuperscript{1,3}, \quad Junyan Lyu\textsuperscript{1,4},
\quad Yijin Huang \textsuperscript{1,5}, \\ Wenhan Luo \textsuperscript{6}, \quad Xiaoying Tang \textsuperscript{1,2} \\
\textsuperscript{1}Department of Electronic and Electrical Engineering,  Southern University of Science and Technology  \\ 
\textsuperscript{2}Jiaxing Research Institute, Southern University of Science and Technology \\
\textsuperscript{3}Department of Electrical and Electronic Engineering, The University of Hong Kong \\
\textsuperscript{4}Queensland Brain Institute, The Univeristy of Queensland \\
\textsuperscript{5}School of Biomedical Engineering,
University of British Columbia \\
\textsuperscript{6}Shenzhen Campus of Sun Yat-sen University \\
}
\maketitle

\begin{abstract}
Contrastive learning based vision-language joint pre-training has emerged as a successful representation learning strategy. In this paper, we present a prototype representation learning framework incorporating both global and local alignment between medical images and reports. In contrast to standard global multi-modality alignment methods, we employ a local alignment module for fine-grained representation. Furthermore, a cross-modality conditional reconstruction module is designed to interchange information across modalities in the training phase by reconstructing masked images and reports. For reconstructing long reports, a sentence-wise prototype memory bank is constructed, enabling the network to focus on low-level localized visual and high-level clinical linguistic features. Additionally, a non-auto-regressive generation paradigm is proposed for reconstructing non-sequential reports. Experimental results on five downstream tasks, including supervised classification, zero-shot classification, image-to-text retrieval, semantic segmentation, and object detection, show the proposed method outperforms other state-of-the-art methods across multiple datasets and under different dataset size settings. The code is available at \url{https://github.com/QtacierP/PRIOR}.
\end{abstract}

\begin{figure*}[t]
  \begin{center}
  \includegraphics[width=0.9\textwidth]{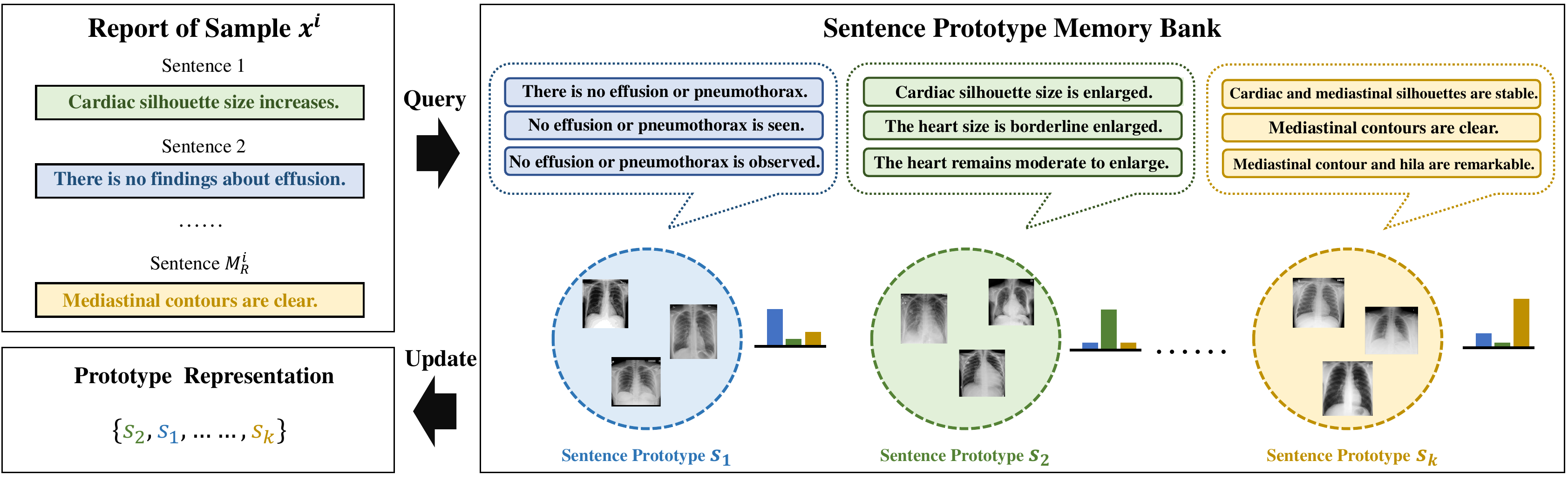}
  \end{center}
  \vspace*{-3mm}
     \caption{Illustration of the proposed sentence-wise prototype memory bank. The prototype embedding can group sentences sharing similar information. Each sentence representation is updated to the nearest prototype after querying.}
  \label{fig:1}
  \vspace*{-5mm}
  \end{figure*}

\section{Introduction}
\label{sec:intro}
Powered by large-scale labeled natural image datasets, deep learning has achieved great success in computer vision \cite{he2016deep, dosovitskiy2020image, ren2015faster, ronneberger2015u, wang2022aesthetic}. However, annotating medical images is extremely expensive and labor-intensive \cite{chen2019self, taleb20203d}. A practical approach is to first pre-train a model on a large-scale labeled natural image dataset like ImageNet \cite{deng2009imagenet}, and then fine-tune it on the downstream medical image dataset with limited annotation 
\cite{raghu2019transfusion, pathak2020deep, zhuang2020comprehensive}. This approach may nevertheless fail to achieve generalized performance due to the domain gap between natural images and medical images \cite{sowrirajan2021moco, ngiam2018domain}. To effectively inherit representation from images of the same domain, self-supervised learning (SSL) methods \cite{chen2020simple, he2020momentum, chen2020improved, chen2021empirical} have been proposed through pre-training on unlabeled datasets. However, it has been suggested that the performance gain of pre-training on unlabeled medical images is relatively limited compared to ImageNet initialization \cite{zhang2020contrastive}. There are potentially two reasons: 1) The sample size of medical images, even unlabeled, is still quite limited compared to that of the ImageNet dataset; 2) Medical images often exhibit high inter-class similarity.

Recently, vision-language pre-training (VLP) has shown natural language supervision can effectively transfer linguistic information to visual representation via well-designed proxy tasks such as contrastive learning \cite{radford2021learning, li2022grounded, yao2021filip, mu2022slip, li2022blip} and generative reconstruction \cite{lu2019vilbert, yu2022coca}. VLP may particularly work for medical image analysis since medical reports are highly likely to be accessible in most situations. However, jointly pre-training medical images and reports is still challenging. First, there are typically multiple sentences in a medical report, and the textual information is highly complex \cite{yang2021writing, li2018hybrid, li2019knowledge}. Second, most descriptions in a medical report are exclusively related to specific sub-regions in the corresponding medical image \cite{muller2021joint}. Most existing VLP methods tend to ignore fine-grained representation and may fail to transfer to locality-aware downstream tasks such as semantic segmentation and object detection. Several methods have employed local alignment losses through contrastive tasks \cite{huang2021gloria, muller2021joint, wang2022multi}. For example, Huang \etal design a localized feature representation framework, but it still falls into the global alignment category \cite{huang2021gloria}. Wang \etal propose a local contrastive loss to align locality-aware information \cite{wang2022multi}, but a vanilla contrastive loss may easily ignore the similarity among nearby sub-regions due to their spatial locations and the overlapped sliding windows in convolution. To address this issue, Muller \etal employ positiveness probability sampling to avoid selecting a nearby sub-region as the negative sample \cite{muller2021joint}. However, this strategy is computationally heavy and time-consuming. Moreover, since contrastive learning is a discriminative SSL paradigm mainly focusing on high-level features, all those aforementioned methods tend to overlook low-level features, such as lesion boundaries in images and symptom descriptions in reports,  which are nevertheless highly crucial for downstream medical image analysis tasks.

In this paper, we present a \textbf{P}rototype \textbf{R}epresentation framework via joint global and local alignment between medical \textbf{I}mages and rep\textbf{OR}ts (PRIOR), wherein we effectively combine contrastive learning and cross-modality conditional reconstruction. We consider sentence in reports and sub-region in images as the elementary local representation units, and the global representation is obtained via attention pooling over localized features. We propose a cross-modality alignment module to align representation between images and reports from both global and local views. To further learn locality-aware and fine-grained information, we utilize an encoder-decoder architecture to maximize the conditional mutual information between paired images and reports. The reconstruction decoder aims to reconstruct the masked image given the report and generate the report's representation given the image. We make use of the prior information that descriptions of most medical reports essentially can be summarized by multiple structured labels \cite{irvin2019chexpert}. That is, sentence-level feature representation can be approximated by prototype categorization without any need to accurately retain redundant information such as syntax. As shown in Figure \ref{fig:1}, each sentence in a report is discretely embedded as prototype representation. In this way, the sentence-level representation learning process can be treated as a classification-like task. Different from the image caption task which predicts each word auto-regressively \cite{wang2022end, liu2021cptr}, sentences in a medical report are usually non-sequential.  Inspired by \cite{carion2020end}, we use a parallel decoder to reconstruct sentence prototype embedding via bipartite matching. We successfully demonstrate the effectiveness of PRIOR on three challenging datasets for five downstream tasks, including supervised classification, zero-shot classification, image-to-text retrieval, semantic segmentation, and object detection. 

Our main contributions are four-fold:
\begin{itemize}
  \item[•] We propose a novel cross-modality alignment module utilizing both global and local information to capture fine-grained features. Compared with previous works, our proposed alignment is more effectively operated on local representation.
  
  \item[•] Considering the writing paradigm of a medical report, we present a prototype memory bank for report's sentence embeddings. Such discrete representation guides the linguistic features to further focus on high-level representation that links tightly to medical images.
  
  \item[•] We leverage a conditional reconstruction task for both vision and language representation, which further facilitates cross-modality feature interaction and explores more structural and causal representation. 
  
  \item[•] PRIOR outperforms existing state-of-the-art (SOTA) methods on five tasks, including supervised classification, zero-shot classification, image-to-text retrieval, semantic segmentation, and object detection.
  \end{itemize}  
  
 \begin{figure*}[t]
  \begin{center}
  \includegraphics[width=0.92\textwidth]{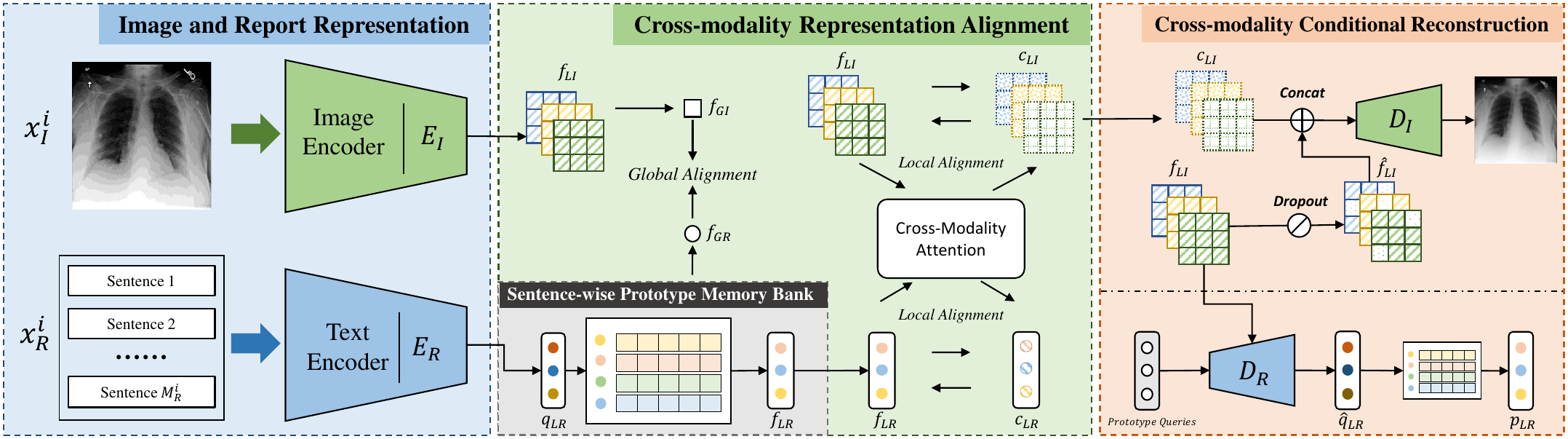}
  \end{center}
  \vspace*{-3mm}
     \caption{The overall framework of the proposed PRIOR. Given a pair of medical image and report, two independent encoders first encode each modality into a common embedding space. Then, the cross-modality alignment module aligns both global and local information between the two modalities. Finally, the cross-modality conditional reconstruction module reconstructs the masked image given the report and generates the sentence prototypes given the image.}
  \label{fig:2}
  \vspace*{-5mm}
  \end{figure*}

\section{Related Work}\label{related work}
{\bf Self-Supervised Learning.} \ \  Self-Supervised Learning (SSL) addresses the dilemma of limited annotated data. There are two main categories: (1) Contrastive SSL methods make use of the invariability of the same image under different augmented views. SimCLR \cite{chen2020simple} and MoCo \cite{he2020momentum, chen2020improved} attempt to maintain the invariability by spreading the representation of different images (negative samples) apart and clustering the representation of the same image (positive sample) together. Recent works have shown specific metric learning frameworks without any negative sample can outperform contrastive learning. Chen \etal propose a simple siamese representation learning framework with low memory and computation cost \cite{chen2021exploring}. Several other works also identify that SSL with no image discrimination could obtain comparable performance to contrastive SSL \cite{grill2020bootstrap}. (2) Generative SSL methods aim to learn data distribution with latent space embedding. Many of them rely on well-designed proxy tasks, such as colorization \cite{larsson2017colorproxy}, image reconstruction \cite{paredes2021back, tewari2018self}, inpainting \cite{pathak2016context}, and so on. Recently, He \etal present a simple masked image auto-encoder with SOTA performance by using vision transformers \cite{he2022masked}.

{\bf Vision-Language Pre-training.} \ \  Vision-Language Pre-training (VLP) is a specific type of SSL, wherein the self-supervised task usually relies on cross-modality interactions between visual and linguistic information. One of the most representative methods is CLIP \cite{radford2021learning}. It shows that contrastive learning can transfer rich semantic information from language supervision to visual representation~\cite{fang2022transferring, chen2019weakly}. However, CLIP only learns coarse-grained representation and ignores localized information. Recently, many works have tried to build fine-grained relationships between natural images and text captions. For example, GLIP converts object detection into a phrase grounding task \cite{li2022grounded}. Yao \etal propose a VLP framework named FLIP involving fine-grained interactions, leveraging localized representation between visual and textual tokens \cite{yao2021filip}. However, none of them are specifically designed for pre-training medical images and reports. Furthermore, GLIP needs annotated bounding boxes for local alignment. And FLIP only focuses on the local alignment between coarse-grained sub-regions and words, which tends to ignore low-level information like textual structure in images and sentence-wise relation in reports. To capture low-level representation, CoCa adopts an encoder-decoder architecture for both contrastive learning and conditional generation \cite{yu2022coca}, but it still cannot deal with the complex structures in medical reports and the low-level information in medical images.

{\bf Medical Image and Report Pre-training.} \ \ In recent years, models that jointly pre-train vision and language for medical utility have been explored. ConVIRT performs global alignment through a contrastive loss \cite{zhang2020contrastive}. On top of ConVIRT, GLoRIA \cite{huang2021gloria}, LoVT \cite{muller2021joint} and MGCA \cite{wang2022multi} all propose their own local alignment mechanisms. GLoRIA and MGCA consider each keyword in a report as the elementary local representation unit, but they do not consider any context across sentences. LoVT aligns sub-regions at the sentence level. However, it still abandons low-level features such as the shape of lesions and detailed symptom descriptions. Inspired by CoCa, we address the issues of existing medical image and report pre-training methods by introducing a conditional reconstruction task combined with prototype representation learning.

{\bf Prototype Representation Learning.} \ \ The goal of prototype representation learning is to cluster similar units into a single embedding. Van \etal show that images could be represented discretely \cite{van2017neural}. And Ramesh \etal prove that prototype learning works well on cross-modality interaction tasks \cite{ramesh2022hierarchical}. Here, we apply prototype learning to medical reports' sentence representation, converting a continuous sentence embedding space into a categorical distribution.

\section{Methodology}
VLP aims to learn a joint distribution $P(X_I, X_R)$ over a group of medical image $X_I=\{x_I^1, ..., x_I^N\}$ and report $X_R=\{x_R^1, ..., x_R^N\}$ pairs. Each sample $x^i$ consists of an image $x_I^i$ and the corresponding report $x_R^i$.

\subsection{Framework Overview}
The overall framework is shown in Figure \ref{fig:2}. Given a pair of an image $x_I^i$ and a report $x_R^i$, we first input them into two separate encoders. Then, each sentence embedding is updated by a learnable sentence-wise prototype memory bank (SPB) for the final linguistic representation. 
In addition to standard global contrastive learning, a local alignment module (LAM) is introduced to align local representation between sub-regions and sentences, aided by a cross-modality attention mechanism. Unlike the common Softmax based attention module, our proposed Sigmoid based LAM makes use of the common sense that not all sentences nor sub-regions are meaningful for cross-modality interaction.  A cross-modality conditional reconstruction (CCR) module is designed to further leverage fine-grained representation. We reconstruct the masked image and the sentence prototypes as guided by the cross-modality representation.

\subsection{Image and Report Representation}
The image encoder $E_I$ encodes each image ${x_I^i}$ into $M_I$ sub-regional representations, formulated as $f_{LI}^{i, v} \in \mathbb{R}^{M_I \times C_I}$, where $v$ is the spatial location index of each sub-region. The global image representation $f_{GI}^{i}$ is obtained by self-attention pooling over localized features \cite{radford2021learning}.

Similarly, the text encoder $E_R$ first encodes each report ${x_R^i}$ into token-wise representations. After that, self-attention pooling is employed to derive $M_R^i$ sentence-wise representations $q_{LR}^{i, u}$ over all tokens in the same sentence, where $u$ indexes sentences and $M_R^i$ is the total number of sentences in the report. Note that all these localized features will be updated by a sentence-wise memory bank that will be described in the next subsection. Different from previous works that consider the [CLS] token as the global representation, we gather sentence-wise features via an additional self-attention pooling operation to serve as the global representation $f_{GR}^i$.

\subsection{Sentence Prototype Memory Bank} \label{spb}
Normally, each sentence in a medical report identifies an observation, including symptoms, locations of lesions, etc. As such, a medical report's sentences can be treated as different representations of prototype features. Therefore, before going through the cross-modality LAM, we update the report's representation via a learnable memory bank. The proposed SPB $S \in \mathbb{R}^{K \times D}$ consists of $K$ representation prototypes, where $D$ is the dimension of the common embedding space. The operation of querying the sentence-wise prototype representation is formulated as
\begin{equation} 
f_{LR}^{i, u} = s_k, \ \ \   k = \mathop{\arg\max}\limits_{j} \frac{q_{LR}^{i, u} \cdot s_j^T}{\| q_{LR}^{i, u}  \| \| s_j \|},
\label {eq:1}
\end{equation}
where $q_{LR}^{i, u}$ is the $u$-th sentence-wise representation of the $i$-th report, and $s_j$ denotes the $j$-th prototype in the memory bank. We update the sentence-wise representation through querying SPB. Since Eq.~\eqref{eq:1} is discrete and not differentiable, we employ the Gumbel-softmax reparameterization trick \cite{jang2016categorical} based on the querying distribution $p_K$ over $K$ cells. To push the localized representation towards the central prototype, an L1 loss is applied to explicitly update the memory bank, namely
\begin{equation} 
 \mathcal{L}_{proto} = \frac{1}{\sum_{i=1}^{N} M_R^i} \sum_{i=1}^{N} \sum_{u=1}^{M_R^i} \| f_{LR}^{i, u} -  q_{LR}^{i, u}\|_1.
\label {eq:2}
  \end{equation}

\subsection{Cross-modality Representation Alignment}
Paired image and report usually describe similar semantic information. For further cross-modality alignment, both global and local representation is projected into a common embedding space with dimension $D$ via four independent MLPs. To explicitly align the global representation between the paired image and report, we maximize their mutual information. Specifically, we employ the InfoNCE loss \cite{oord2018representation} to estimate the lower bound of the mutual information. The global report-to-image alignment loss is defined as
\begin{equation} 
\mathcal{L}_{g}^{I\gets R} = -\frac{1}{B} \sum_{i=1}^B\log \frac{\exp(f_{GI}^i \cdot {f_{GR}^{i}}^T / \tau_1)}{\sum_{j=1}^B \exp(f_{GI}^i \cdot {f_{GR}^{j}}^T  / \tau_1)}, 
\end{equation}
where $B$ is the batch size and $\tau_1$ is the temperature parameter.  Similarly, the global image-to-report alignment loss is formulated as
\begin{equation} 
 \mathcal{L}_{g}^{R\gets I} = -\frac{1}{B} \sum_{i=1}^B\log \frac{\exp(f_{GI}^i \cdot {f_{GR}^{i}}^T / \tau_1)}{\sum_{j=1}^B \exp(f_{GI}^j \cdot {f_{GR}^{i}}^T  / \tau_1)}.
\end{equation}

Global alignment focuses on discriminative high-level features. However, medical image analysis tasks are generally highly sensitive to fine-grained low-level information like lesion boundaries. As such, we design an LAM. 

The first step of LAM is to generate cross-modality attention representation. Specifically, sub-region/sentence representation is respectively learned by sub-region-wise/sentence-wise weighted sum. Given that not all sub-regions/sentences contain meaningful semantic information, we do not use the Softmax function which regards each localized unit as contributing constantly to the cross-modality representation (\textit{i.e.}, sum of weights is 1). The local report-to-image attention-based representation is formulated as
\begin{equation} 
c_{LI}^{i, v}  = \sum_{k=1}^{M_R^i}\sigma \left(\frac{Q^If_{LI}^{i, v}   \cdot {K^If_{LR}^{i, k}}^T}{\sqrt{D}} \right) \cdot V^If_{LR}^{i, k},
\label{eq:attention}
\end{equation}
where $Q^I$, $K^I$ and $V^I$ are learnable matrices, and $\sigma$ denotes the Sigmoid function. Similarly, the local image-to-report attention-based representation is 
\begin{equation} 
  c_{LR}^{i, u} = \sum_{k=1}^{M_I}\sigma \left(\frac{Q^Rf_{LR}^{i, u}   \cdot {K^Rf_{LI}^{i, k}}^T}{\sqrt{D}} \right) \cdot V^Rf_{LI}^{i, k}.
\end{equation}

After obtaining the cross-modality attention representation, we perform local alignment for localized cross-modality information interaction. For local image-to-report alignment, the cross-modality representation and the localized linguistic representation of the same sentence are regarded as a positive pair, while those from different sentences are negative samples. We observe that different reports may share similar sentence representation. Therefore, for local alignment we use only sentences from the same report as the negative samples. As such, the image-to-report local alignment loss is defined as
\begin{equation} 
\mathcal{L}_{l}^{R\gets I} = -\frac{1}{\sum\limits_{i=1}^{B} M_R^i} \sum_{i=1}^{B}\sum\limits_{u=1}^{M_R^i}\log \frac{\exp(f_{LR}^{i, u} \cdot {c_{LR}^{i, u}}^T / \tau_2)}{\sum_{k=1}^{M_R^i} \exp(f_{LR}^{i,u} \cdot {c_{LR}^{i,k}}^T  / \tau_2)},
\end{equation}
where $\tau_2$ is the temperature parameter. For local report-to-image alignment, we conjecture that nearby sub-regions contain similar information induced by spatial structure and overlapping convolutions. It may cause feature collapse if we consider a positive-like sample as a negative one. Therefore, we apply the alignment methods proposed in \cite{chen2021exploring}, utilizing neither negative pairs in contrastive loss nor prior sampling strategies in \cite{muller2021joint}. The report-to-image local alignment loss is based on the cosine similarity and gets combined with asymmetrical projection, namely
\begin{equation} 
\begin{split}
 \mathcal{L}_{l}^{I\gets R} &= -\frac{1}{B M_I}\sum_{i=1}^{B}\sum_{v=1}^{M_I} \frac{1}{2} \text{\bf{sim}}\left(h(f_{LI}^{i, v}), S(c_{LI}^{i, v})\right) \\ &+ \frac{1}{2} \text{\bf{sim}}\left(h(c_{LI}^{i, v}), S(f_{LI}^{i, v})\right),
\end{split} 
\end{equation}
where $h$ is an independent MLP head, $S$ is the stop-gradient operation, and $\text{\bf{sim}}$ denotes the cosine similarity. Finally, the total cross-modality representation alignment loss is
\begin{equation} 
  \mathcal{L}_{align} = \mathcal{L}_{g}^{I\gets R} + \mathcal{L}_{g}^{R\gets I}  + \mathcal{L}_{l}^{I\gets R} + \mathcal{L}_{l}^{R\gets I}.
\end{equation}

\subsection{Cross-modality Conditional Reconstruction}
Both global and local alignments concentrate on high-level features but are likely to ignore low-level information. To address this issue, we introduce a cross-modality conditional reconstruction task aiming to recover low-level information based on cross-modality interaction. 

We consider two reconstruction tasks: masked image reconstruction and sentence prototype generation. We randomly zero out 50\% image features via dropout. Then we concatenate the masked image features $\hat{f}_{LI}^{i}$ and the unmasked cross-modality features $c_{LI}^{i}$. To reconstruct the image, a lightweight decoder $D_I$ is applied. The reconstruction loss of the masked image is presented as
\begin{equation} 
  \mathcal{L}_{ir} = \frac{1}{N M_I}\sum_{i=1}^{N} \| D_I\left(\text{\bf{cat}}(\hat{f}_{LI}^{i}, c_{LI}^{i})\right)  - x_I^i\|_1,
\end{equation}
where $\text{\bf{cat}}$ is the concatenation operation, $\hat{f}_{LI}^{i}$ represents the masked feature map, and $c_{LI}^{i}$ denotes the cross-modality feature map.

For reconstructing the report information, we follow an encoder-decoder architecture and apply a cross-attention mechanism \cite{vaswani2017attention} to interact with visual features. Inspired by \cite{carion2020end}, parallel decoding is used to generate the sentence-wise prototypes. We employ $U$ learnable embeddings (we call prototype queries) to generate different predicted prototypes in the decoder. If the number of sentences $M_R^i$ is smaller than $U$, we pad the sentence embeddings with zero prototype $\varnothing$. Since the sentences in a report are not necessarily sequential, which means the decoding process should be permutation invariant, we apply the Hungarian algorithm \cite {kuhn1955hungarian} to identify the optimal prediction-label pair $\hat{\sigma}$ in the matching space $\mathfrak{S}_U$. We consider the L1 distance as the matching cost to perform bipartite matching
\begin{equation} 
\hat{\sigma} = \mathop{\arg\min}\limits_{\sigma \in \mathfrak{S}_U} \sum_{i=0}^{N}  \sum_{u=0}^{M_R^i} \| s^{i, u} - \hat{q}_{\sigma(i, u)} \|_1,
\end{equation}
where $s^{i, u}$ is the ground truth prototype and $\hat{q}_{\sigma(i, u)}$ is the predicted querying embedding that matches index $\sigma(i, u)$. After deriving the optimal matching pairs, we apply Eq.~\eqref{eq:1} to obtain the predicted prototype $\hat{s}_{\sigma(i)}$ based on the predicted distribution $q_K$. 

The loss function for sentence prototype prediction has three components. The first term is the L1 distance between the predicted querying embedding and the target prototype
\begin{equation} 
\mathcal{L}_q = \frac{1}{\sum_{i=1}^{N} M_R^i}  \sum_{i=0}^{N}\sum_{u=0}^{M_R^i} \| s^{i, u} - \hat{q}_{\sigma(i, u)} \|_1.
\end{equation}
The second term focuses on distribution consistency, namely the KL divergence between the querying distribution $p_K$ and the predicted distribution $q_K$
\begin{equation} 
  \mathcal{L}_{kl} = D_{KL}(q_k \| p_k).
\end{equation}
The last term is similar to that in the global alignment. We first obtain the predicted global report representation $p_{GR}^{i}$ via self-attention pooling over predicted prototypes $p_{LR}$. Then a global prediction alignment loss aiming to align the predicted global representation and the realistic global report representation is formulated as follows
\begin{equation} 
  \mathcal{L}_{gpa} = -\sum_{i=1}^B\log \frac{\exp(f_{GR}^i \cdot {p_{GR}^{i}}^T / \tau_3)}{\sum_{j=1}^B \exp(f_{GR}^j \cdot {p_{GR}^{i}}^T  / \tau_3)}.
\end{equation}
Finally, the overall loss for the cross-modality conditional reconstruction is
\begin{equation} 
  \mathcal{L}_{recon} = \mathcal{L}_{ir} + \mathcal{L}_q  + \mathcal{L}_{kl} + \mathcal{L}_{gpa}.
\end{equation}

\subsection{Loss}
The overall loss function of the proposed PRIOR contains the following terms: the cross-modality representation alignment loss $\mathcal{L}_{align}$, the SPB updating loss $\mathcal{L}_{proto}$ and the cross-modality conditional reconstruction loss $\mathcal{L}_{recon}$. We optimize these terms jointly via a weighted sum
\begin{equation} 
  \mathcal{L} = \mathcal{L}_{align} +  \lambda_{proto} \mathcal{L}_{proto} + \lambda_{recon} \mathcal{L}_{recon},
\end{equation}
where $\lambda_{proto}$ and $\lambda_{recon}$  are hyper-parameters.

\begin{table*}[t]
\setlength{\abovecaptionskip}{0.5cm}
\caption{Supervised classification results by fine-tuning on three downstream datasets. All methods are trained on different portions of the training set from 1\% to 100\% and evaluated by AUC-ROC. Each value is the average of five runs. The best results are highlighted in bold and red, and the second-best results are highlighted in blue.}
  \centering
   \vspace{2mm}
  \resizebox{0.9\textwidth}{!}{%
  \begin{tabular}{@{}lccccccccccc@{}}
  \toprule
\multirow{2}{*}{Init. Methods}                     & \multicolumn{3}{c}{RSNA}  &  & \multicolumn{3}{c}{SIIM} &  & \multicolumn{3}{c}{CheXpert} \\ \cmidrule(r){2-4} \cmidrule(lr){6-8} \cmidrule(l){10-12} 
  \multicolumn{1}{l}{} &
    \multicolumn{1}{c}{1\%} &
    \multicolumn{1}{c}{10\%} &
    \multicolumn{1}{c}{100\%} &
     &
    \multicolumn{1}{c}{1\%} &
    \multicolumn{1}{c}{10\%} &
    \multicolumn{1}{c}{100\%} &
     &
    \multicolumn{1}{c}{1\%} &
    \multicolumn{1}{c}{10\%} &
    \multicolumn{1}{c}{100\%} \\ \midrule
  Random Init.                         & 56.42 $\pm$ 2.79 & 57.62 $\pm$ 3.27 &                   83.63 $\pm$ 0.27  &  &    55.41 $\pm$ 1.72    &   57.83 $\pm$ 0.60     &   80.63 $\pm$ 0.55       &  &     52.96 $\pm$ 3.52     &     73.37 $\pm$  0.68   &    75.12 $\pm$ 0.35     \\
  ImageNet Init.                       & 80.27 $\pm$ 0.71 & 81.85 $\pm$ 0.46 &  88.01 $\pm$ 0.13                   &  &     78.33 $\pm$ 1.79   &   82.88 $\pm$ 0.37     &  88.41 $\pm$ 0.41      &  &       77.57 $\pm$ 1.13   &    82.94 $\pm$  0.91    &     87.08 $\pm$ 0.37   \\ \midrule
  MoCo        \cite{he2020momentum}                        & 82.33 $\pm$ 0.47 & 85.22 $\pm$ 0.11 & 87.90 $\pm$ 0.09                    &  &  75.49 $\pm$  0.29      &  81.01 $\pm$  0.67      &   88.43 $\pm$ 0.24     &  &      78.00 $\pm$ 0.62    &      86.27 $\pm$ 0.30   &  87.24 $\pm$     0.05   \\
  MoCoV2     \cite{chen2020improved}                         & 83.07 $\pm$ 0.49 & 85.88  $\pm$ 0.28 &  88.60 $\pm$ 0.07                   &  &   77.10 $\pm$  0.49   &   81.12 $\pm$ 0.64    &   90.72 $\pm$ 0.21   &  &     79.64 $\pm$ 0.53     &      86.04 $\pm$ 0.23   &    87.44 $\pm$ 0.27    \\
  SimCLR           \cite{chen2020simple}                   & 80.18 $\pm$ 2.78   & 84.60 $\pm$ 0.12 &    88.07 $\pm$ 0.11                 &  &    74.97 $\pm$ 2.17    &    83.21 $\pm$ 0.49    &    88.72 $\pm$ 0.28    &  &        67.41 $\pm$ 2.74  &  86.74 $\pm$ 0.36    &  87.97 $\pm$ 0.22       \\ \midrule
  ConVIRT        \cite{zhang2020contrastive}                     & 84.17 $\pm$ 0.77  & 86.92 $\pm$ 0.13  &          88.74 $\pm$  0.36          &  &     84.17 $\pm$ 0.77   &     85.66 $\pm$ 0.45   &   91.50 $\pm$ 0.08     &  &      85.02 $\pm$ 0.28    &       87.58 $\pm$  0.53     &     88.21 $\pm$ 0.46    \\
  GLoRIA     \cite{huang2021gloria}                         &84.12 $\pm$  0.47 & 86.83 $\pm$ 0.53 & 89.13 $\pm$ 0.12                    &  &   85.05 $\pm$ 1.62     &   88.51 $\pm$ 0.78     &   92.11 $\pm$ 0.18     &  &       83.61 $\pm$ 0.52   &   87.40 $\pm$ 0.39      &  88.34 $\pm$ 0.12      \\
  BioVil \cite{boecking2022making}    & 81.95 $\pm$ 0.29  & 85.37 $\pm$ 0.61 & 88.62 $\pm$ 0.14  &  & 79.89 $\pm$ 1.06 &  81.62 $\pm$ 2.37 & 90.48 $\pm$ 0.14 &  & 80.77 $\pm$ 1.26  &  87.56 $\pm$ 0.16   &  \textcolor{blue}{88.41 $\pm$ 0.51}    \\
 LoVT \cite{muller2021joint}                    & 85.51 $\pm$ 0.36 & 86.53 $\pm$ 0.54 &      \textcolor{blue}{89.27 $\pm$ 0.13}              &  &   85.47 $\pm$ 0.85     & 88.50 $\pm$ 0.76       & \textcolor{blue}{92.16 $\pm$ 0.36}       &  &   85.13 $\pm$ 0.48       &   \textcolor{blue}{88.05 $\pm$ 0.15}      & 88.27 $\pm$ 0.07       \\
 MGCA         \cite{wang2022multi}                    & \textcolor{red}{\bf{85.80 $\pm$ 0.68}} & \textcolor{red}{\bf{87.66 $\pm$ 0.21}} & \textcolor{red}{\bf{89.30 $\pm$ 0.16}}  &  & \textcolor{blue}{86.10 $\pm$ 0.31}  & \textcolor{red}{\bf{89.68 $\pm$ 0.09}} & 92.04 $\pm$ 0.09 &  &    \textcolor{blue}{85.63 $\pm$ 0.33}  &   87.65 $\pm$ 0.33  &  88.30 $\pm$ 1.48    \\
 \textbf{PRIOR (ours)} &\textcolor{blue}{85.74 $\pm$ 0.36} & \textcolor{blue}{87.08 $\pm$ 0.19} & 89.22 $\pm$ 0.17 & & \textcolor{red}{\bf{87.27 $\pm$ 0.39}} & \textcolor{blue}{89.13 $\pm$ 0.11}  &\textcolor{red}{\bf{92.39 $\pm$ 0.23}} &  & \textcolor{red}{\bf{86.16 $\pm$ 0.64}}  & \textcolor{red}{\bf{88.31 $\pm$ 0.20}} &  \textcolor{red}{\bf{88.61 $\pm$ 0.29}}       \\ \bottomrule
  \end{tabular}%
  }
     \label{Tab:supervised-cls}
  \vspace{-2mm}
  \end{table*}

\section{Experiments}
\subsection{Datasets}
{\bf MIMIC-CXR} \cite{johnson2019mimic}. \ \ We pre-train all SSL and VLP methods on version 2 of MIMIC-CXR, which contains $377110$ X-ray chest images and $227827$ corresponding reports. Because downstream tasks all employ frontal-view images, we remove all lateral-view images.  Additionally, we abandon short reports containing fewer than four sentences, ending up with $182475$ image-report pairs. 

{\bf RSNA Pneumonia Detection} \cite{shih2019augmenting}. \ \ The RSNA Pneumonia Detection dataset contains more than $30000$ X-ray chest images. All data have image-level pneumonia classification and bounding-box pneumonia detection labels. Therefore, we utilize the RSNA dataset for two downstream tasks: supervised classification and object detection. For the classification task, we split the dataset by 60\%/20\%/20\% for training/validation/testing. For the object detection task, we only use the pneumonia samples (a total of 6012) through five-fold cross-validation.

{\bf SIIM-ACR Pneumothorax}. \ \ SIIM-ACR Pneumothorax contains $12954$ X-ray chest images, together with image-level pneumothorax annotation and pixel-level segmentation mask if pneumothorax exists. We use them for downstream supervised classification and semantic segmentation. For the classification task, we split the dataset by 60\%/20\%/20\% for training/validation/testing. For the segmentation task, we only use the pneumothorax samples (a total of 2669) through five-fold cross-validation.

{\bf CheXpert}. \cite{irvin2019chexpert} \ \  CheXpert contains more than $220000$ X-ray chest images. Each image is labeled for five independent diseases. And thus the downstream task for this dataset is multi-label classification. We split the training set by 80\%/20\% for training/validation and use the official validation set for testing. Similar to that in GLoRIA \cite{huang2021gloria}, we create a sub-set (named CheXpert 5x200) from the training set of CheXpert for zero-shot classification and image-to-text retrieval, including $200$ exclusively positive samples from each disease.

\subsection{Implementation Details}
Following previous works, we use the BERT model pre-trained by ClinicalBERT \cite{alsentzer2019publicly}. For image feature encoding, we apply ResNet50 \cite{he2016deep} as the backbone. We resize all images into $224 \times 224$, and all  reports are padded or truncated to have a fixed length of 256 tokens. We choose 768 to be the common latent dimension $D$. The SPB size $K$ is set as $512$. The temperature coefficients $\tau_1$, $\tau_2$, and $\tau_3$ are all set as $0.01$. The loss weight coefficients are set as $\lambda_{recon}=1$ and $\lambda_{proto}=10$. The overall framework is trained with a batch size of $128$ for $100$ epochs. The learning rate is initialized as $1e-5$ and decays following a cosine policy. 

\subsection{Evaluation Experiments}
We compare our method with several widely-used initialization methods, including three image-level SSL methods (MoCo, MoCoV2, SimCLR), five medical VLP methods (ConVIRT, GLoRIA, BioVil, LoVT, MGCA), and ImageNet pre-training. All SSL and VLP methods are re-trained on the MIMIC-CXR dataset with their official implementations and default hyper-parameters. ConVIRT applies only a global alignment loss, and may not perform well on locality-aware downstream tasks. GLoRIA, LoVT and MGCA are three SOTA VLP methods targeting medical images and reports with well-designed local alignment losses. 

\begin{figure}[]
  \vspace*{-4mm}
  \begin{center}
  \includegraphics[width=0.46\textwidth]{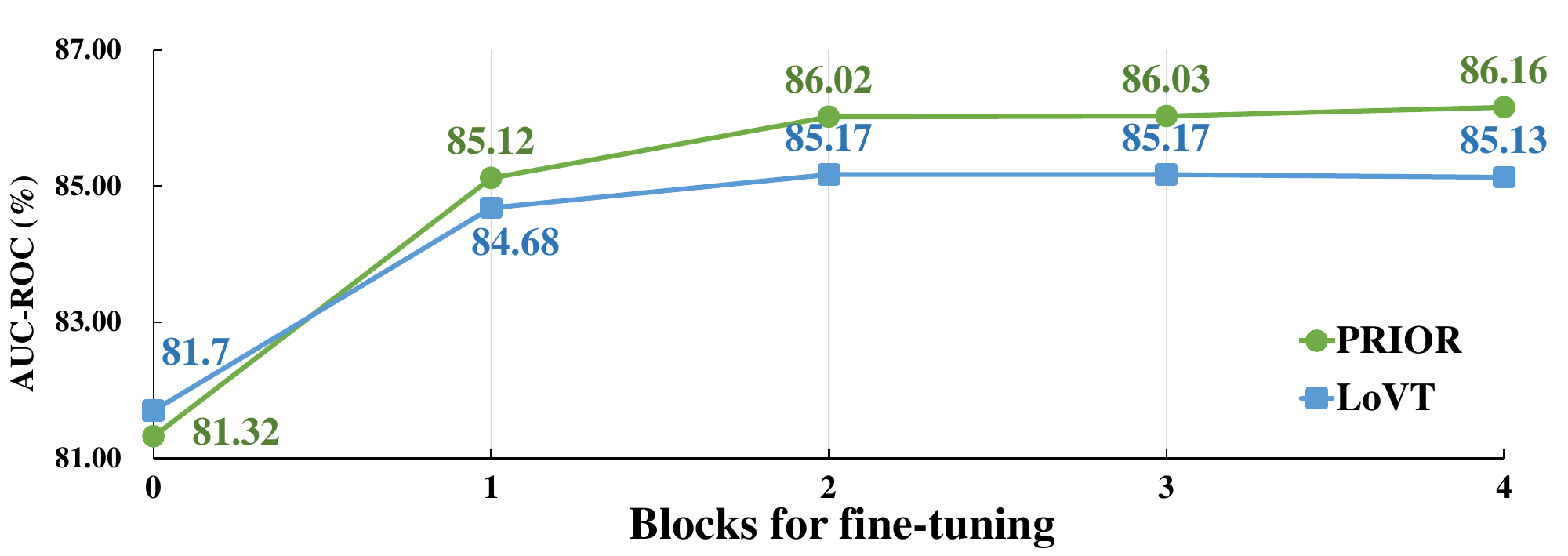}
  \end{center}
  \vspace*{-4mm}
     \caption{Partial fine-tuning results on 1\% CheXpert. The number of blocks for fine-tuning increases from left to right. Fine-tuning with 0 block is equivalent to linear evaluation, while fine-tuning with 4 blocks is equivalent to full fine-tuning.}
\label{fig:finetune}
\vspace*{-5mm}
\end{figure}

All compared methods are evaluated on five downstream tasks, including supervised classification, zero-shot classification, image-to-text retrieval, semantic segmentation, and object detection. For zero-shot classification and image-to-text retrieval, we compare with GLoRIA and ConVIRT which present prompt engineering in their original papers. Although most of the existing methods perform linear evaluation, similarly to \cite{he2022masked}, we conduct a partial fine-tuning evaluation on CheXpert with 1\% training data, updating only the last several residual blocks in ResNet50. We observe that the linear evaluation performance is not strictly keeping in line with the fine-tuning performance, as clearly shown in Figure \ref{fig:finetune}. Additionally, if we unfreeze only the last block, the performance significantly increases, which agrees with previously-reported results \cite{he2022masked}. Linear evaluation cannot well deal with non-linear features which are nevertheless critical for downstream tasks. Therefore, we only report the fine-tuning results.

\subsection{Results on Classification}
We conduct supervised classification experiments on RSNA, SIIM, and CheXpert respectively. AUC-ROC is employed as the evaluation metric. Table \ref{Tab:supervised-cls} shows that all VLP methods outperform image-level SSL methods and ImageNet initialization. Collectively, PRIOR achieves the best results, especially on SIIM and CheXpert with different training data sizes. Even though ConVIRT, GLoRIA, BioVil and LoVT have leveraged semantic representation from large-scale report supervision, they may have ignored the clinical paradigm relation between images and reports, the relation of which is successfully modeled by our proposed SPB. MGCA conducts disease-level alignment and performs well on the RSNA dataset, but it neglects sentence-level clinical information which is nevertheless crucial for complex downstream tasks like multi-label classification on CheXpert.

To analyze the quality of the cross-modality representation, we measure the zero-shot classification performance on CheXpert 5x200. We use models pre-trained on MIMIC-CXR to predict the labels of CheXpert. We compare PRIOR with two VLP methods, namely ConVIRT and GLoRIA, on the zero-shot classification task. For generalization, we employ the simplest prompt engineering strategy without any heavy ad-hoc design. The prediction is given by measuring the similarly between the predicted image and prompts. We take the sum of the global alignment loss and the cross-modality local attention as the similarity. As shown in Table \ref{Tab:zero}, neither ConVIRT nor GLoRIA can capture correlative semantic information with simple prompt engineering. PRIOR attains the best zero-shot classification results, suggesting it can effectively transfer knowledge from linguistic supervision to vision domain.

\subsection{Image-to-text Retrieval}
We conduct image-to-text retrieval experiments on the CheXpert 5x200 dataset. Since the reports in CheXpert are inaccessible, we randomly select 1000 reports from MIMIC-CXR with exclusive 200 samples for each of 5 diseases. We measure the performance using Precision@K to check if the retrieved report matches the query image label. Our results in Table \ref{Tab:retrieval} demonstrate that PRIOR significantly outperforms both GLoRIA and ConVIRT.

\subsection{Results on Segmentation and Detection}
We examine the contribution of the local representation through the semantic segmentation task. We use U-Net \cite{ronneberger2015u} as the segmentation architecture. Table \ref{Tab:segmentation} presents the results of all methods on the SIIM dataset. Since GLoRIA, LoVT, MGCA and the proposed PRIOR consider localized feature representation during the pre-training phase, all of them outperform other methods. Among all the four medical image and report VLP methods, PRIOR achieves the most superior performance, especially with limited training data. A plausible reason is that PRIOR learns detailed locality-aware information via conditional reconstruction. 

\begin{table}[]
\caption{Zero-shot classification results on CheXpert 5x200.}
  \vspace{2mm}
  \centering
  \resizebox{0.4\textwidth}{!}{%
  \begin{tabular}{lccc}
  \toprule
  \multirow{2}{*}{VLP Methods} & \multicolumn{3}{c}{CheXpert Zero-shot Classification} \\ \cline{2-4} 
                  & Accuracy & F1-Score & Precision\\ \hline
  ConVIRT  \cite{zhang2020contrastive}       &  21.30   &  \textcolor{blue}{19.03}         &  17.65       \\
  GLoRIA    \cite{huang2021gloria}      &  \textcolor{blue}{23.20}   &   15.79       &      \textcolor{red}{\bf{44.01}}  \\ 
  \textbf{PRIOR (ours)} &  \textcolor{red}{\bf{34.90}}   &   \textcolor{red}{\bf{30.56}}   &     \textcolor{blue}{35.88}   \\ \bottomrule
  \end{tabular}%
  }
  \vspace{-2mm}
  \label{Tab:zero}
  \end{table}

\begin{table}[t]
\caption{Image-to-text retrieval results on CheXpert 5x200.}
  \centering
  \vspace{2mm}
  \resizebox{0.4\textwidth}{!}{%
  \begin{tabular}{lcccc}
    \toprule
  \multirow{2}{*}{VLP Methods} & \multicolumn{4}{c}{CheXpert Image-to-text Retrieval} \\ \cline{2-5} 
                                & Prec $@$ 1        & Prec $@$ 2       & Prec $@$ 5  & Prec $@$ 10   \\ \hline
  ConVIRT                \cite{zhang2020contrastive}      &     20.3                 &      19.8                &  19.7        &       19.9               \\
 GLoRIA                \cite{huang2021gloria}      &     \textcolor{blue}{29.3}                 &      \textcolor{blue}{29.0}                &  \textcolor{blue}{27.8}        &       \textcolor{blue}{26.8}               \\
 \textbf{PRIOR(ours)}   &      \textcolor{red}{\bf{40.2}}                &      \textcolor{red}{\bf{39.6}}                &   \textcolor{red}{\bf{39.3}}       &   \textcolor{red}{ \bf{38.0}}         \\ \bottomrule
  \end{tabular}%
  }
\vspace{-2mm}
  \label{Tab:retrieval}
  \end{table}

To further evaluate the effectiveness of our proposed method, we conduct object detection on the RSNA dataset. We use Faster R-CNN \cite{ren2015faster} as the detection architecture. Table \ref{Tab:detection} presents the performance comparison results on RSNA. We observe that some SSL methods outperform VLP methods, and we conjecture it is because medical reports may be incapable of precisely providing pneumonia information, which is consistent with the original motivation of building the RSNA dataset \cite{shih2019augmenting}. With that being said, our proposed PRIOR still outperforms all other methods, successfully establishing its effectiveness.

\begin{table}[]
\caption{Semantic segmentation results by fine-tuning on SIIM. All methods are trained on different portions of the training set from 1\% to 100\% and evaluated by Dice.}
   \vspace{2mm}
  \centering
  \resizebox{0.4\textwidth}{!}{%
 \begin{tabular}{lccc}
  \toprule
  \multirow{2}{*}{Init. Methods} & \multicolumn{3}{c}{SIIM Segmentation} \\ \cline{2-4} 
                                & 1\%       & 10\%       & 100\%       \\ \hline
  Random Init.                   &  6.59 $\pm$ 0.40        &    16.61 $\pm$ 1.44        &  38.25 $\pm$ 0.90           \\
  ImageNet Init.                 &     14.92 $\pm$  3.38    &    25.26 $\pm$ 3.04 
  &     45.07 $\pm$ 0.87
  \\ \hline
  MoCo        \cite{he2020momentum}                  &    17.66 $\pm$  2.14       &      27.81 $\pm$ 2.19      &    41.18 $\pm$ 1.53           \\
  MoCoV2             \cite{chen2020improved}           &    18.19 $\pm$ 3.35     &    28.57 $\pm$ 2.92        &   44.01 $\pm$ 1.36        \\
  SimCLR              \cite{chen2020simple}          &     16.55 $\pm$ 2.27      &    23.36 $\pm$ 0.66        &        40.62 $\pm$ 0.69     \\ \hline
  ConVIRT      \cite{zhang2020contrastive}                 &      18.48 $\pm$ 2.43     &     27.32 $\pm$ 2.60
  &     41.72 $\pm$ 0.69
  \\
  GLoRIA       \cite{huang2021gloria}                 &     18.78 $\pm$ 2.85      &  \textcolor{blue}{33.55 $\pm$ 2.14}          & 45.46 $\pm$ 1.69            \\
  BioVil        \cite{boecking2022making}                  &   18.13 $\pm$ 1.67    &     27.78 $\pm$ 1.10       &        \textcolor{blue}{45.54 $\pm$ 1.14}    \\
  LoVT        \cite{muller2021joint}                  &    18.81  $\pm$ 1.04      &     32.68 $\pm$ 0.95       &        44.65 $\pm$ 1.36     \\
  MGCA     \cite{wang2022multi}                 &       \textcolor{blue}{18.84 $\pm$ 2.37}    &       33.54 $\pm$ 1.89   &    45.39 $\pm$ 1.35        \\
  \textbf{PRIOR (ours)}                         &  \textcolor{red}{\bf{20.43 $\pm$ 2.20}}         &     \textcolor{red}{\bf{34.81 $\pm$ 1.60}}       &    \textcolor{red}{\bf{46.01 $\pm$ 1.03}}          \\ \bottomrule
  
  \end{tabular}%
  }
   \label{Tab:segmentation}
  \end{table}

\subsection{Ablation Study}
This section presents the ablation study results of PRIOR under the 1\% SIIM dataset setting for supervised classification and semantic segmentation. We compare the performance of the following variants: (1) Baseline, which only uses global alignment; (2) Baseline+LAM, which uses both global alignment and local alignment; (3) Baseline+LAM+CCR, which uses global alignment, local alignment and cross-modality conditional reconstruction; (4) Baseline+LAM+CCR+SPB (proposed PRIOR), which uses global alignment, local alignment, cross-modality conditional reconstruction, and sentence prototype memory bank. Note that variant (3) reconstructs only the masked image because the sentence-level representation is not categorized and unavailable without SPB. Table \ref{Tab:ablation} tabulates the ablation study results, demonstrating all components are essential for extracting both global and local features.

\begin{figure}[]
  \begin{center}
  \includegraphics[width=0.45\textwidth]{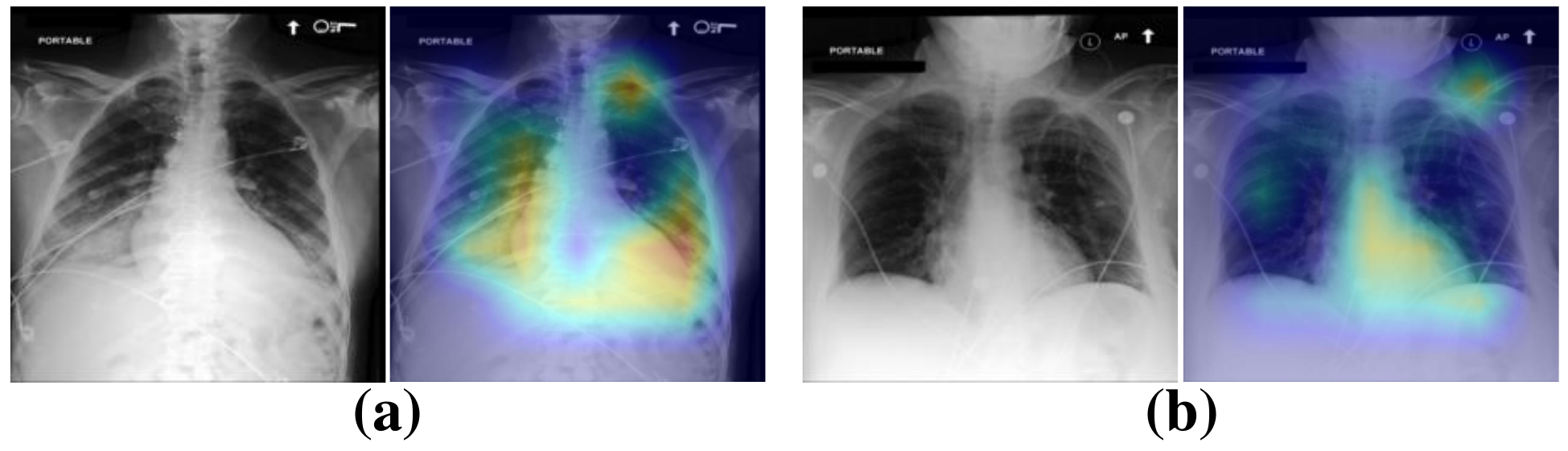}
  \end{center}
  \vspace{-5mm}
     \caption{Representative cross-modality attention maps. (a) The related sentence is ``\textit{increased bibasilar opacities are combination of increased bilateral pleural effusions and bibasilar atelectasis}''. (b) The related sentence is ``\textit{unchanged normal size of the cardiac silhouette}
     ''.}
\label{fig:attention}
\vspace{-5mm}
\end{figure}

\subsection{Visualization of Cross-modality Attention}
We visualize the cross-modality attention map to show the effectiveness and interpretability of the proposed PRIOR. Given an image and a corresponding sentence, we first obtain the cross-modality attention map by Eq. \eqref{eq:attention}. Then, we visualize the attention map by resizing it to the same size as the image and superimposing it on top of the image. Figure \ref{fig:attention} presents two examples to illustrate that the proposed attention module can detect correlative regions given a sentence. For example, the heatmap in Figure \hyperref[fig:attention]{4a} highlights the abnormal regions precisely, aligning well with the report. Figure \hyperref[fig:attention]{4b} also correctly detects the heart location.

\subsection{Visualization of Feature Representation}
To assess the quality of the feature representation from the proposed method, we utilize t-SNE \cite{van2008visualizing} to visualize the high-level embeddings from the last layer of the image encoder on CheXpert 5x200. Apparently, results from the proposed method exhibit a clustering pattern more evidently, while results from other methods are more isotropic and homogeneous and fail to distinguish different categories. 

\begin{table}[t]
    \caption{Object detection results by fine-tuning on RSNA. All methods are trained on different portions of the training set from 1\% to 100\% and evaluated by mAP(0.5:0.95). \textbf{-} means the method's mAP is lower than 0.1.}
  \centering
  \vspace{2mm}
  \resizebox{0.4\textwidth}{!}{%
  \begin{tabular}{lccc}
  \toprule
  \multirow{2}{*}{Init. Methods} & \multicolumn{3}{c}{RSNA Object Detection} \\ \cline{2-4} 
                                & 1\%       & 10\%       & 100\%       \\ \hline
  Random Init.                   &   \bf{-}         &     2.66 $\pm$ 1.34       &    18.99 $\pm$ 1.36         \\
  ImageNet Init.                 &    \bf{-}        &       7.85 $\pm$ 2.68     &        21.16 $\pm$ 1.94     \\ \hline
  MoCo \cite{he2020momentum}                    &    \bf{-}        &         16.25 $\pm$ 1.96   &       21.00 $\pm$ 2.00     \\
  MoCoV2        \cite{chen2020improved}                &    \textcolor{blue}{0.19 $\pm$ 0.22}        &     18.45 $\pm$  1.75      &   21.65 $\pm$ 2.10          \\
  SimCLR              \cite{chen2020simple}          &      \bf{-}      &       14.00 $\pm$ 1.96     &      21.76 $\pm$ 1.93       \\ \hline
  ConVIRT       \cite{zhang2020contrastive}                &       \bf{-}     &     18.80 $\pm$ 2.79       &  21.28   $\pm$ 1.65         \\
  GLoRIA             \cite{huang2021gloria}           &      \bf{-}     &       17.95 $\pm$  1.37     &        21.37 $\pm$ 1.07     \\
  BioVil        \cite{boecking2022making}                  &   \bf{-}   &    \textcolor{blue}{19.11   $\pm$ 1.13}     &   19.50 $\pm$ 0.97    \\
  LoVT                \cite{muller2021joint}          &       \bf{-}   &     17.46 $\pm$ 2.21    &       \textcolor{blue}{21.80 $\pm$ 2.75}      \\
  MGCA                \cite{wang2022multi}          &       \bf{-}   &     19.10 $\pm$ 1.83    &       21.33 $\pm$ 1.63      \\
  \textbf{PRIOR (ours)}                        & \textcolor{red}{\bf{0.20 $\pm$   0.27}}       &     \textcolor{red}{\bf{19.61 $\pm$ 1.93}}      &    \textcolor{red}{\bf{22.20 $\pm$ 1.58}}         \\ \bottomrule
  \end{tabular}%
  }
\vspace{-2mm}
  \label{Tab:detection}
  \end{table}

\begin{table}[t]
  \caption{Ablation results of the proposed method on 1\% SIIM. LAM denotes the local alignment module, CCR denotes the cross-modality conditional reconstruction and SPB denotes the sentence prototype memory bank.}
  \vspace{2mm}
  \centering
  \resizebox{0.4\textwidth}{!}{%
  \begin{tabular}{@{}ccccc@{}}
  \toprule
  \multicolumn{3}{c}{Components} & \multirow{2}{*}{1\% SIIM Classification} & \multirow{2}{*}{1\% SIIM Segmentation} \\ \cmidrule(r){1-3}
  LAM & CCR & SPB &  &  \\ \midrule
     &   &     & 82.01 $\pm$ 2.21 &   17.88 $\pm$ 1.24\\ 
  \checkmark   &   &    & 84.86 $\pm$ 0.65 & 18.70 $\pm$ 2.33 \\
  \checkmark    &  \checkmark   &    & 85.80 $\pm$ 0.86  & 19.33  $\pm$ 2.22\\
  \checkmark    &     &  \checkmark  &  86.81 $\pm$ 0.47 &  19.87 $\pm$ 2.03\\
  \checkmark    &  \checkmark   &  \checkmark    &\bf{87.27 $\pm$ 0.39} & \bf{20.43 $\pm$ 2.20} \\ \bottomrule
  \end{tabular}%
  }
\vspace{-2mm}
  \label{Tab:ablation}
  \end{table}

  \begin{figure}[]
    \begin{center}
    \includegraphics[width=0.45\textwidth]{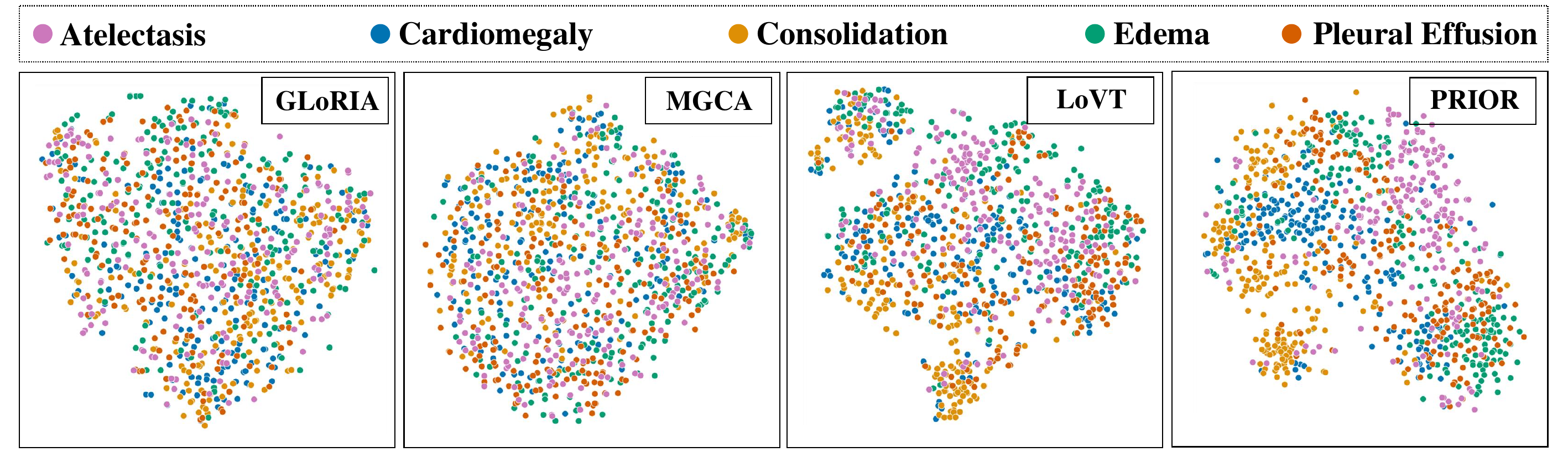}
    \end{center}
    \vspace*{-4mm}
       \caption{t-SNE visualization of the high-level embeddings from the last layer of the image encoder on CheXpert 5x200.}
  \label{fig:TSNE}
  \vspace*{-5mm}
  \end{figure}

\section{Conclusion}
In this paper, we have explored fine-grained representation learning from paired medical images and reports. We present a cross-modality alignment strategy by learning from both global and local information. A sentence-wise prototype memory bank is proposed to translate clinical descriptions into categorical representations and cluster similar clinical conditions. Based on the prototype memory bank, a cross-modality conditional reconstruction module is designed to reconstruct masked images and missing sentence prototypes through cross-modality interaction, facilitating low-level representation learning which is crucial for locality-aware downstream tasks. Extensive experiments on five downstream tasks demonstrate the effectiveness of the proposed method.

\section{Acknowledgements}
This study was supported by the Shenzhen Basic Research Program
(JCYJ20190809120205578); the National Natural Science Foundation
of China (62071210); the Shenzhen Science and Technology Program
(RCYX20210609103056042); the Shenzhen Basic Research Program
(JCYJ20200925153847004); the Shenzhen Science and Technology Innovation Committee (KCXFZ2020122117340001).

{\small
\bibliographystyle{ieee_fullname}
\bibliography{egbib}
}

\clearpage

\begin{large} 
\textbf{\quad \quad \quad \quad  Supplementary Material}
\end{large}

\section{Introduction}
In this document, we provide supplementary material for our paper ``PRIOR: Prototype Representation Joint Learning from Medical Images and Reports''. We first provide pre-training details. Then we present details of fine-tuning on different downstream tasks. Finally, we perform detailed analysis of each key component in PRIOR, including sentence-wise prototype memory bank (SPB), local alignment module (LAM), and cross-modality conditional reconstruction (CCR).

\section{Pre-training Details}
\subsection{Data Preprocessing}
We pre-train the proposed PRIOR on the MIMIC-CXR-JPG dataset. We remove all lateral-view images and images whose corresponding reports have fewer than four sentences. Finally, we end up with $182475$ image-report pairs. 

For image preprocessing, we first normalize the intensities of all images into $[0, 1]$ and then calculate their mean ($0.4755$) and standard deviation ($0.3011$). Z-score normalization is employed. All images are resized to $224 \times 224$ before being fed into a model of interest. 

For report preprocessing, we utilize all sentences from the \textit{Findings} and \textit{Impression} sections.  We employ BioClinicalBERT's tokenizer \cite{alsentzer2019publicly} implemented in the Transformer library \cite{wolf-etal-2020-transformers} to tokenize each report into a sequence of tokens. We then pad all reports to have the same length of $256$ tokens.

\subsection{Single-modality Encoder Architecture}
We employ ResNet50 as the image encoder, which is implemented in the TorchVision library \cite{marcel2010torchvision} and gets pre-trained on ImageNet. For global representation, we apply an attention pooling layer to obtain a $2048$-dimensional vector. For local representation, we take the feature map $f \in \mathbb{R}^{7 \times 7 \times 2048}$ of the last layer of ResNet50.

We adopt BioClinicalBERT \cite{alsentzer2019publicly} as the report encoder, which is implemented in the Transformer library \cite{alsentzer2019publicly} and gets pre-trained on the MIMIC-CXR dataset. We take the hidden state from the last layer as the token-level representation. After that, we gather all token representations in the same sentence via self-attention pooling to serve as the sentence-level representation $q_i \in \mathbb{R}^{M_R^i \times 768}$, where $M_R^i$ is the number of sentences in the $i$-th report. Finally, we use an additional self-attention pooling layer to obtain the global representation over all sentence-level representations. 

To embed linguist and visual representation into the same dimension, we attach four independent MLPs to embed both local and global report/image representation into an embedding space of dimension $768$.

\subsection{Cross-modality Interaction Architecture}
The cross-modality interaction architecture mainly contains three components, namely sentence-wise prototype memory bank, cross-modality alignment, and cross-modality conditional reconstruction.

{\bf Sentence-wise prototype memory bank.} \ Each prototype in SPB is initialized with the Standard Gaussian distribution and L1 normalization. For stable convergence, the temperature coefficient in Gumbel-softmax reparameterization decays from $0.9$ to $0.01$.

{\bf Cross-modality alignment.} \  We employ the contrastive loss implemented in PyTorch-Lightning \cite{falcon2019pytorch} as the global alignment loss and the local image-to-report alignment loss. We adopt the loss proposed elsewhere \cite{chen2021exploring} as the local report-to-image alignment loss. 

{\bf Cross-modality conditional reconstruction.} \ We adopt five up-sampling blocks for image reconstruction. Each block consists of two convolution operations with a size of $3 \times 3$ and a stride of $1 \times 1$, followed by a ReLU activation function and a batch normalization layer.  For report reconstruction, we adopt a decoder with 6 BERT layers in the same setting as that of BioClinicalBERT \cite{alsentzer2019publicly}. We employ the bipartite matching algorithm to match the predicted prototypes with the ground-truth ones, which is implemented in the SciPy library \cite{2020SciPy-NMeth}.

\subsection{Training Details}
We train PRIOR for 100 epochs on NVIDIA A100 GPUs with a batch size of 128. The total training process consists of three stages: (1) In stage 1, we train the cross-modality alignment module for 20 epochs, which is directly trained on sentence-wise representation without SPB. (2) In stage 2, we jointly train SPB and the cross-modality alignment module for 30 epochs. (3) In stage 3, we jointly train SPB, the cross-modality alignment module, and the cross-modality conditional reconstruction module for 50 epochs.  We use the Adam optimizer \cite{kingma2014adam} with a learning rate of $1e-5$ and a weight decay of $1e-6$.  A cosine annealing scheduler is employed to adjust the learning rate. We employ grid search to identify the optimal hyper-parameter combination that performs the best on downstream tasks. 

\section{Fine-tuning Details}
We evaluate the proposed PRIOR on five downstream tasks, namely supervised classification, zero-shot classification, image-to-text retrieval, semantic segmentation, and object detection. For each task, we employ grid search to identify the best-performing hyper-parameters. We here only provide the specific hyper-parameters for VLP methods since they are our essential objects of interest; all VLP methods share the same set of hyper-parameters. For hyper-parameters employed in non-VLP methods, please refer to our source code for details. The intensity of each image is normalized into $[0, 1]$ via Z-score normalization using the mean and standard deviation of the training set. We resize all images into $224 \times 224$. A cosine decay scheduler is used to adjust the learning rate.  

\subsection{Supervised Classification}
For the downstream task of supervised classification, we fine-tune the image encoder with an additional fully-connected layer on the RSNA dataset, the SIIM Pneumothorax dataset, and the CheXpert dataset. The hyper-parameters of all VLP methods are listed in Table \ref{Tab:hyper-parameters-cls}. Since each dataset involves either binary classification or multi-label classification, we use the binary cross-entropy loss as the loss function. The model with the highest AUC-ROC on the validation set is selected for testing.

\begin{table}[]
    \caption{Hyper-parameter details for supervised classification.}
  \vspace{2mm}
  \centering
  \resizebox{0.4\textwidth}{!}{%
    \begin{tabular}{ccccc}
    \hline
    Dataset &
    \begin{tabular}[c]{@{}c@{}}Training Data \\ Ratio\end{tabular} &
    Learning Rate &
    Epochs &
    Batch Size \\ \hline
    \multirow{3}{*}{\begin{tabular}[c]{@{}c@{}}RSNA \end{tabular}} &
      \multicolumn{1}{c|}{100\%}  &    $1e-4$           &   5     &  64
       \\
     & \multicolumn{1}{c|}{10\%} &        $1e-4$       &   5     &  64  \\
     & \multicolumn{1}{c|}{1\%}   &  $1e-4$            &    10    &  64  \\ \hline \hline
    \multirow{3}{*}{\begin{tabular}[c]{@{}c@{}}SIIM\end{tabular}} &
      \multicolumn{1}{c|}{100\%} &  $1e-5$             &   5     &   64
       \\
     & \multicolumn{1}{c|}{10\%}   &     $1e-5$          &    20    &  64  \\
     & \multicolumn{1}{c|}{1\%}   &       $1e-4$        &    20    &      64  \\ \hline \hline
     \multirow{3}{*}{\begin{tabular}[c]{@{}c@{}}CheXpert\end{tabular}} &
     \multicolumn{1}{c|}{100\%}  &        $1e-6$       &   5     &  256
      \\
    & \multicolumn{1}{c|}{10\%} &      $1e-5$         &    5    &  256  \\
    & \multicolumn{1}{c|}{1\%}   &     $1e-4$          &  5      &   256  \\ \hline
    \end{tabular}%
    }
  \vspace{-4mm}
  \label{Tab:hyper-parameters-cls}
  \end{table}

\subsection{Zero-shot Classification}
We conduct zero-shot classification on the CheXpert 5x200 dataset, which is based on the similarity between manually designed prompt sentences and images. We generate the prompt simply through using ``\textit{X is observed}'', where \textit{X} represents the name of a specific disease. For generalization, we do not use any ensemble method. We calculate the similarity between the prompt and the image via the sum of the global alignment loss and the local cross attention weight. Note that before summing the two scores, we employ the softmax function to normalize each of them into $[0, 1]$.

\subsection{Image-to-text Retrieval}
Similar to zero-shot classification, we conduct image-to-text retrieval on the CheXpert 5x200 dataset, which is based on the similarity between queried reports and images. Note that because CheXpert does not have public reports, we sample 1000 exclusive reports from MIMIC-CXR dataset for each of 5 diseases.

\subsection{Semantic Segmentation}
We adopt U-Net \cite{ronneberger2015u} as the segmentation network. Different from the original U-Net, we employ U-Net equipped with a ResNet50 backbone, which is implemented in the Segmentation Models Pytorch library \cite{Iakubovskii:2019}. We present the hyper-parameter details in Table \ref{Tab:hyper-parameters-finegrained}. The Dice loss is used for training, which performs better than either the cross-entropy loss or a combination of the two losses according to our experimental results. The model with the highest Dice score on the validation set is selected for testing.

\subsection{Object Detection}
We adopt Faster-RCNN \cite{ren2015faster} as the object detection network. The training and evaluation scripts are based on Pytorch-Lightning Flash \cite{falcon2019pytorch}. The hyper-parameter details are also tabulated in Table \ref{Tab:hyper-parameters-finegrained}. We follow the original Faster-RCNN setting in terms of the loss function \cite{ren2015faster}. The model with the highest mAP (0.5:0.95) on the validation set is selected for testing.

\begin{table}[]
 \caption{Hyper-parameter details for semantic segmentation and object detection.}
   \vspace{2mm}
  \centering
  \resizebox{0.4\textwidth}{!}{%
  \begin{tabular}{ccccc}
  \hline
  Dataset &
    \begin{tabular}[c]{@{}c@{}}Training Data \\ Ratio\end{tabular} &
    Learning Rate &
    Epochs &
    Batch Size \\ \hline
  \multirow{3}{*}{\begin{tabular}[c]{@{}c@{}}SIIM \\ Segmentation\end{tabular}} &
    \multicolumn{1}{c|}{100\%}  &     $1e-3$          &    50    &    128
     \\
   & \multicolumn{1}{c|}{10\%} & $1e-4$ & 50 & 128  \\
   & \multicolumn{1}{c|}{1\%}  & $1e-4$ & 100 & 16  \\ \hline \hline
  \multirow{3}{*}{\begin{tabular}[c]{@{}c@{}}RSNA\\ Detection\end{tabular}} &
    \multicolumn{1}{c|}{100\%} &      $5e-5$       &    5    &   16
     \\
   & \multicolumn{1}{c|}{10\%}  & $1e-4$ & 20 &  16  \\
   & \multicolumn{1}{c|}{1\%}   & $1e-4$ & 5 & 16  \\ \hline
  \end{tabular}%
  }
  \vspace{-2mm}
  \label{Tab:hyper-parameters-finegrained}
  \end{table}

\section{Component Analysis}
In this section, we perform further analysis on each component of PRIOR, including SPB, LAM, and CCR. 

\subsection{Analysis on SPB}
\begin{figure*}[b!]
    \begin{center}
    \includegraphics[width=1.0\textwidth]{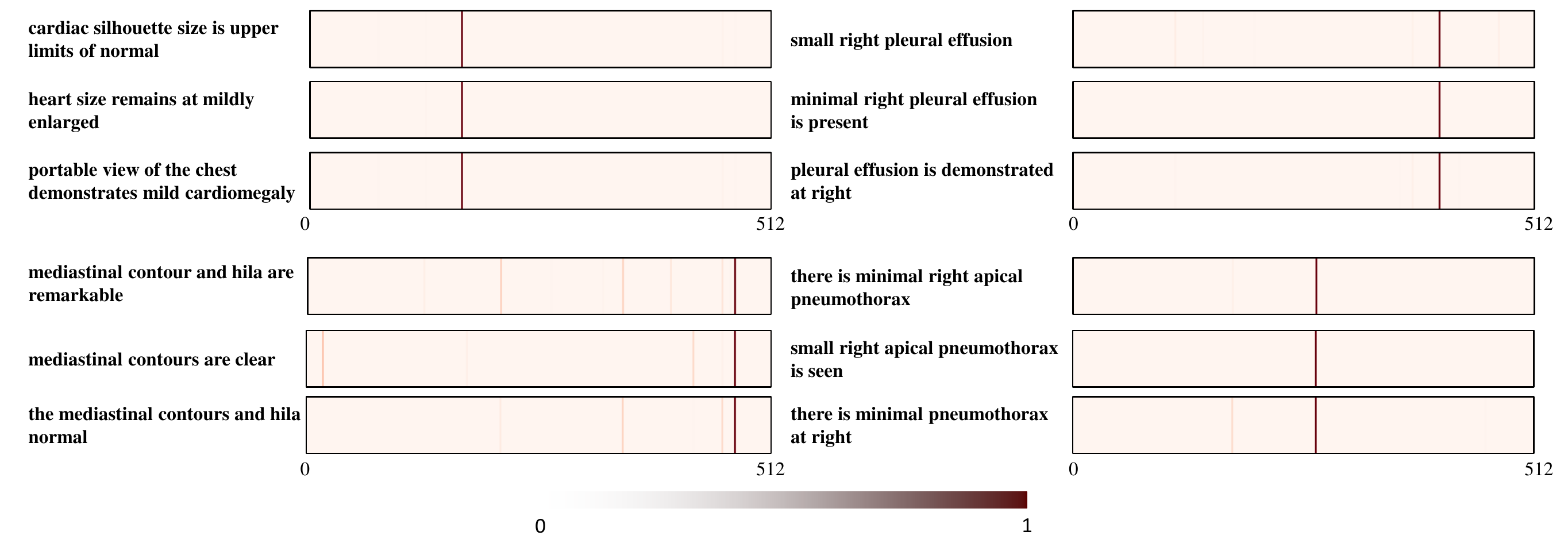}
    \end{center}
    \vspace*{-2mm}
       \caption{Representative examples demonstrating that SPB can cluster sentences with similar information. The horizontal axis represents the sentence index in the memory bank, and the color bar quantifies the querying score. }
    \label{fig:spb}
    \vspace*{4mm}
    \end{figure*}

To demonstrate the effectiveness of SPB, we utilize several synonymous sentences to demonstrate the querying distribution over the memory bank, as shown in Figure \ref{fig:spb}. It is worth noting that SPB highlights the most relevant prototype in a sharp distribution over SPB, which means that SPB can effectively capture the high-level semantic information of the sentence of interest.
  \begin{figure}[]
    \begin{center}
    \includegraphics[width=0.4\textwidth]{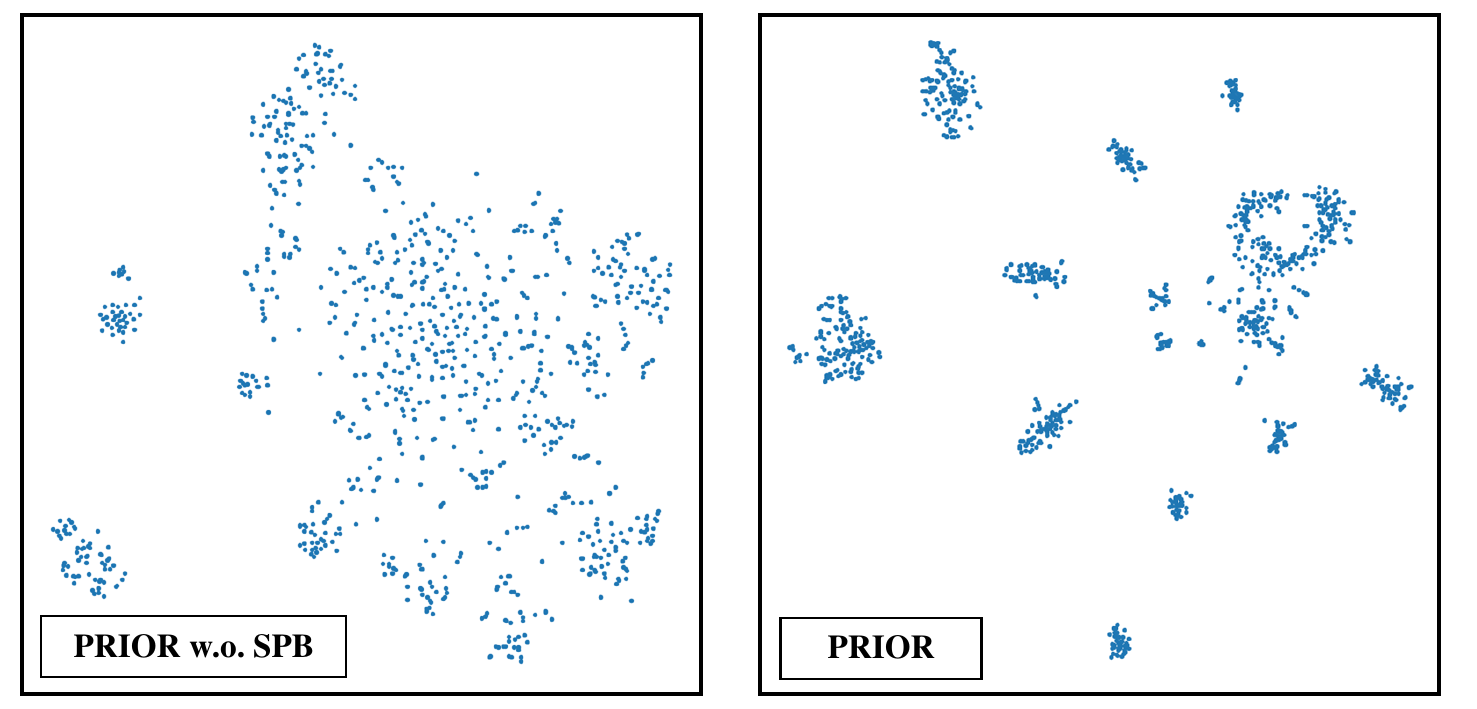}
    \end{center}
    \vspace*{-2mm}
       \caption{The t-SNE visualization comparison between the sentence-level embeddings without and with SPB pre-training.}
    \label{fig:tsne}
    \vspace*{-2mm}
    \end{figure}
    
Furthermore, we visualize the sentence-wise embeddings via t-SNE \cite{van2008visualizing} on 1000 randomly selected sentences from MIMIC-CXR in Figure \ref{fig:tsne}. Apparently, the sentence-wise embeddings with SPB pre-training are more compact and well-separated. In other words, SPB significantly improves the quality of the sentence embeddings, which benefits cross-modality interaction.

\subsection{Analysis on LAM}
\begin{figure*}[b]
    \begin{center}
    \includegraphics[width=1.0\textwidth]{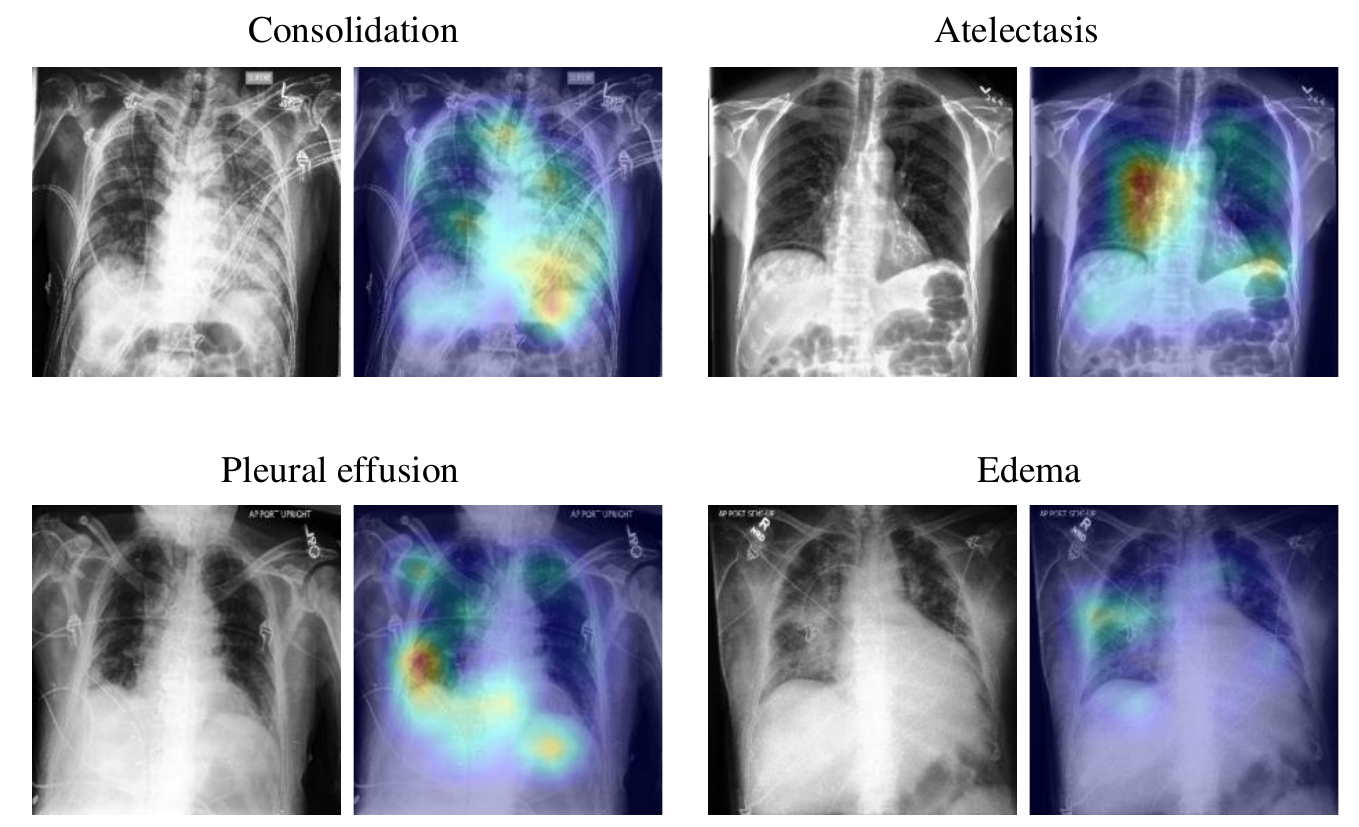}
    \end{center}
    \vspace*{0mm}
       \caption{Representative examples of cross-modality attention maps related to different diseases.}
    \label{fig:lam}
    \vspace*{-7mm}
    \end{figure*}

In Figure \ref{fig:lam}, we visualize the attention weights of LAM on CheXpert 5x200. We first generate a sentence in the form of ``\textit{X is observed}'', where \textit{X} represents the name of a specific disease. Then the cross-modality attention map is obtained by the proposed LAM over the generated sentence and the corresponding image. Clearly, LAM can effectively localize the affected region of the given disease. 

Unlike natural image captioning, some sentences in medical reports are irrelevant to medical images, such as ``AP single view of the chest has been obtained". Softmax assigns a constant probability summing up to 1 when aligning sentence with image, resulting in an average representation of image and failing to represent local features precisely. Instead, Sigmoid assigns probability to each pixel separately. Our experiments demonstrate the effectiveness of the Sigmoid function in cross-modality attention, as shown in Table \ref{Tab:activation}.

\begin{table}[]
\centering
\caption{The influence of the activation functions in LAM.}
\vspace{2mm}
\resizebox{0.4\textwidth}{!}{%
\begin{tabular}{lcc}
\hline
Activation function & CheXpert Classification & SIIM Segmentation \\ \hline
Softmax             & 86.01 $\pm$ 0.68        & 45.65 $\pm$ 1.31  \\
Sigmoid             & \textbf{86.16 $\pm$ 0.64}        & \textbf{46.01 $\pm$ 1.03}  \\ \hline
\end{tabular}%
}
\label{Tab:activation}
\vspace{-1mm}
\end{table}

\subsection{Analysis on CCR}
\begin{figure*}[]
    \begin{center}
    \includegraphics[width=1.0\textwidth]{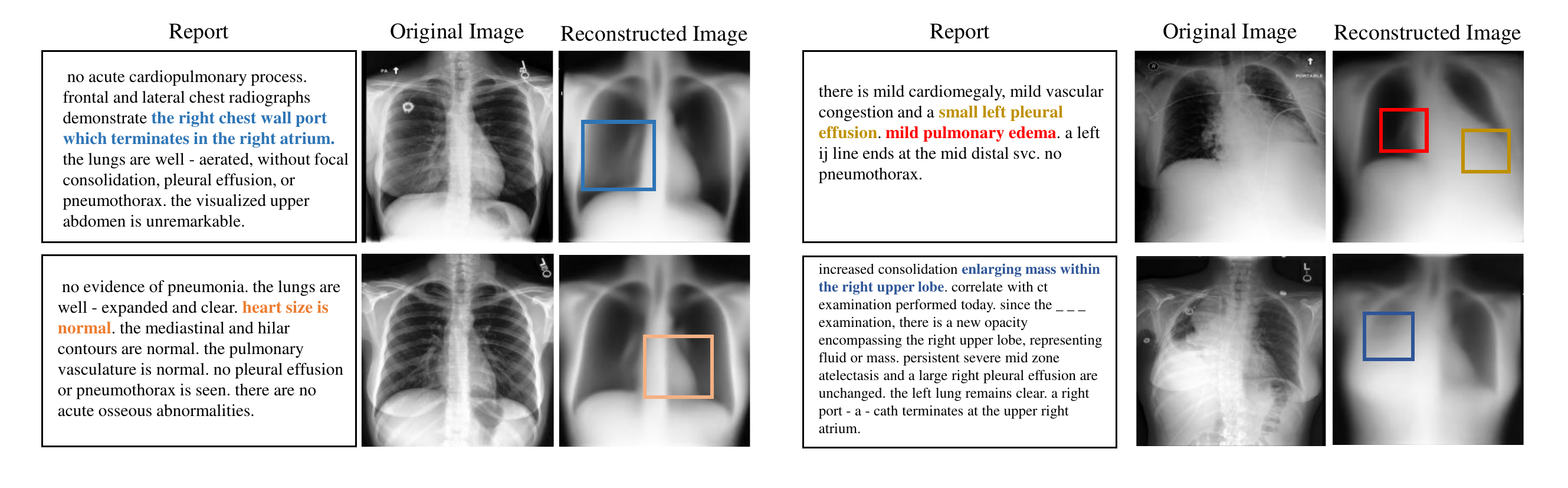}
    \end{center}
    \vspace*{-2mm}
       \caption{Representative examples of reconstructed images from CCR.}
    \label{fig:ir}
    \vspace*{-7mm}
    \end{figure*}

CCR is a key component in our PRIOR, which can effectively capture fine-grained cross-modality information. We visualize representative reconstructed images from the CCR module in Figure \ref {fig:ir}. We find that the reconstructed images can well maintain the low-level information that is related to the report descriptions. Meanwhile, the reconstructed images successfully reveal the severity of the corresponding disease as well as the lesion locations.

\begin{figure*}[t]
      \begin{center}
      \includegraphics[width=1.0\textwidth]{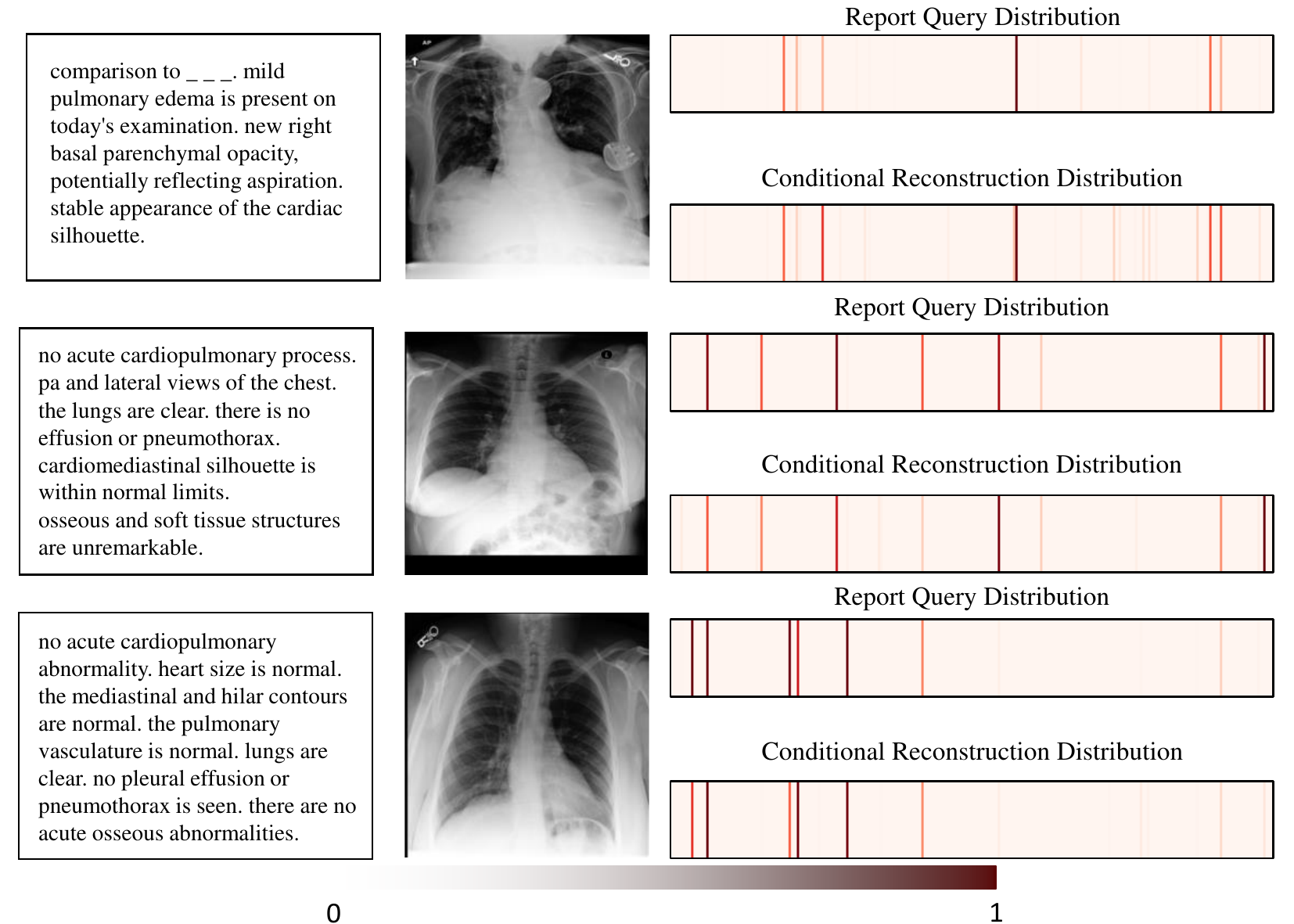}
      \end{center}
      \vspace*{-2mm}
         \caption{Representative examples of conditional reconstruction for sentence-wise prototypes. From left to right, the first column shows the original reports, the second column shows the original images, the top panel of the third column shows report distributions over the memory bank, and the bottom panel of the third column shows the predicted sentence representation distributions. Note that the top and bottom panels of the third column are very similar to each for all provided examples.}
      \label{fig:spg}
      \vspace*{-7mm}
      \end{figure*}

For sentence prototype reconstruction, we compare the query distribution of the original report and the predicted distribution from CCR in Figure \ref{fig:spg}. The distribution is obtained from all sentences in the report of interest. We observe that the predicted distribution is similar to the original distribution, which demonstrates that the CCR module can effectively capture cross-modality information and reconstruct sentence prototypes.

\end{document}